\documentclass[review]{elsarticle}

\usepackage{lineno,hyperref}
\usepackage{amsmath}
\usepackage{amsfonts}
\usepackage{array}
\usepackage{multirow}
\usepackage{booktabs}
\usepackage{url}

\usepackage{graphicx}
\usepackage{amssymb}
\usepackage{booktabs}
\usepackage{enumitem}
\usepackage{bbm}
\usepackage[ruled,linesnumbered]{algorithm2e}
\usepackage{color}
\usepackage[table]{xcolor}

\journal{Expert Systems with Applications}

\begin{document}

\begin{frontmatter}

\title{ME-Mamba: \textbf{M}ulti-\textbf{E}xpert \textbf{Mamba} with Efficient Knowledge Capture and Fusion for Multimodal Survival Analysis}

\author[mymainaddress,mysecondaryaddress]{Chengsheng Zhang}
\ead{cszhang24@m.fudan.edu.cn}
\author[mymainaddress,mysecondaryaddress]{Linhao Qu}
\ead{lhqu20@fudan.edu.cn}
\author[mymainaddress,mysecondaryaddress]{Xiaoyu Liu}
\ead{liuxiaoyu21@m.fudan.edu.cn}
\author[mymainaddress,mysecondaryaddress]{Zhijian Song\corref{mycorrespondingauthor}}
\ead{zjsong@fudan.edu.cn}

% \fntext[firstAuthor]{Linhao Qu and Shaolei Liu contributed equally to this work.}
\cortext[mycorrespondingauthor]{Corresponding author}

\address[mymainaddress]{Digital Medical Research Center, School of Basic Medical Science, Fudan University, Shanghai 200032, China}
\address[mysecondaryaddress]{Shanghai Key Lab of Medical Image Computing and Computer Assisted Intervention}

\begin{abstract}
Survival analysis using whole-slide images (WSIs) is crucial in cancer research. Despite significant successes, pathology images typically only provide slide-level labels, which hinders the learning of discriminative representations from gigapixel WSIs. With the rapid advancement of high-throughput sequencing technologies, multimodal survival analysis integrating pathology images and genomics data has emerged as a promising approach. However, the high dimensionality of the data and the heterogeneity between modalities pose major challenges for extracting discriminative features and effectively fusing modalities. To address these issues, we propose a \textbf{M}ulti-\textbf{E}xpert \textbf{Mamba} (ME-Mamba) system that captures discriminative pathological and genomic features while enabling efficient integration of both modalities. This approach achieves complementary information fusion without losing critical information from individual modalities, thereby facilitating accurate cancer survival analysis. Specifically, we first introduce a Pathology Expert and a Genomics Expert to process unimodal data separately. Both experts are designed with Mamba architectures that incorporate conventional scanning and attention-based scanning mechanisms, allowing them to extract discriminative features from long instance sequences containing substantial redundant or irrelevant information. Second, we design a Synergistic Expert responsible for modality fusion. It explicitly learns token-level local correspondences between the two modalities via Optimal Transport, and implicitly enhances distribution consistency through a global cross-modal fusion loss based on Maximum Mean Discrepancy. The fused feature representations are then passed to a mamba backbone for further integration. Through the collaboration of the Pathology Expert, Genomics Expert, and Synergistic Expert, our method achieves stable and accurate survival analysis with relatively low computational complexity. Extensive experimental results on five datasets in The Cancer Genome Atlas (TCGA) demonstrate our state-of-the-art performance. We will make our code publicly available.
\end{abstract}

\begin{keyword}
survival analysis, Mamba, multimodal learning, multiple instance learning
\end{keyword}

\end{frontmatter}

% \linenumbers

\section{Introduction}
Survival analysis, a core task in clinical prognostic research and cancer outcome assessment, aims to predict the time until the occurrence of an event such as death or disease recurrence from a specific starting point and to accurately evaluate patients’ mortality risk. It plays a vital role in enhancing diagnosis and informing treatment planning within clinical decision-making processes\cite{1,2}. For cancer patients, multimodal data such as pathology images and genomics data provide interrelated and critical information, forming the basis for patient stratification and survival analysis\cite{3}. However, conventional survival analysis methods often rely on short-term clinical indicators and long-term follow-up reports\cite{4,5,6,7}, which are not only time-consuming but also limited in clinical applicability. Meanwhile, the complex nature of cancer necessitates comprehensive assessment of diverse and personalized data, posing significant challenges for models to effectively capture key discriminative features and integrate data heterogeneity. Therefore, developing efficient feature extraction and multimodal fusion methods has become essential yet challenging for building robust and accurate survival analysis models.

Recently, with the rapid advancement of deep learning technologies, significant progress has been made in medical image analysis. As the gold standard for cancer diagnosis, pathology images are increasingly being applied in survival analysis\cite{8,9,10,11,12}. Pathology images directly provide microscopic morphological features of tumor cells and information about the tumor microenvironment. Such visual characteristics are closely associated with tumor progression, invasiveness, and patient prognosis, offering a morphological basis for assessing survival risk. However, using pathology images alone can't capture molecular-level biological information. Many prognostic factors are not directly reflected in morphological features, making it difficult to reveal their deep correlation with survival outcomes through image analysis alone. This may lead to an incomplete interpretation of tumor heterogeneity. Therefore, multimodal survival analysis methods that integrate pathology images and genomics data\cite{13,14,15,16,17,18,19} hold considerable research promise. To accurately perform survival analysis using these two modalities, most existing methods employ Transformer-based architectures to enable cross-modal interaction and obtain multimodal representations. For example, some approaches use Transformer-based multiple instance learning to capture global information\cite{14}, or apply co-attention mechanisms for modality fusion\cite{19}, thereby acquiring complementary information from different perspectives. Although these methods have demonstrated strong performance in feature modeling and cross-modal interaction, high-dimensional multimodal features can easily obscure key survival-related information originally present in unimodal data. This increases the risk of overfitting to task-irrelevant features. Moreover, the quadratic computational complexity of the attention mechanism results in low efficiency when processing long sequences or large-scale multimodal data, often causing the neglect of critical instance-level features.

\begin{figure*}[t!]
    \centering
    \includegraphics[width=\textwidth]{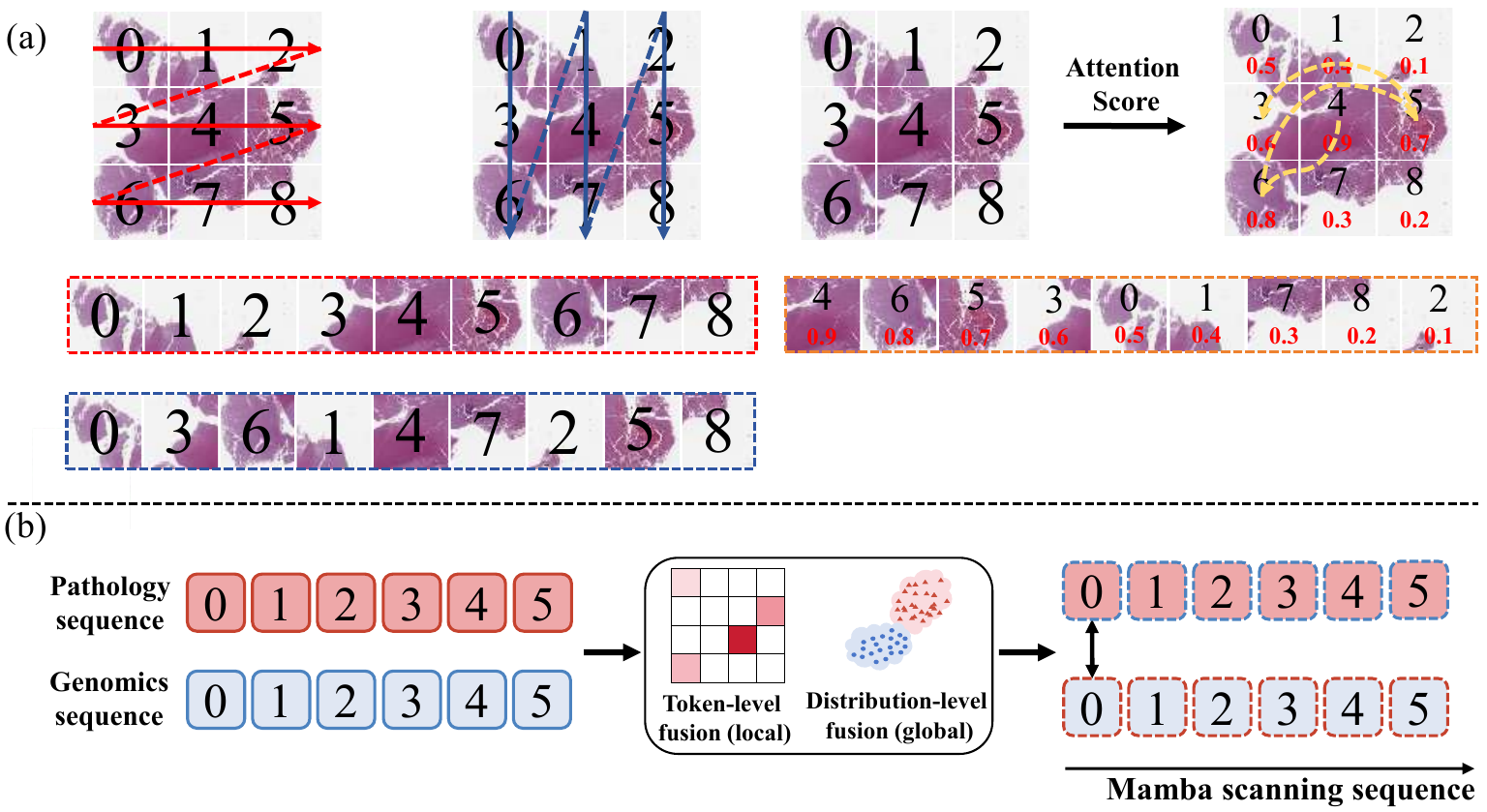}
    \caption{The operational mechanism of each expert module. (a) Illustration of different scanning strategies in unimodal expert. Two conventional scanning strategies directly scan the instance sequence from left to right and from top to bottom, whereas the attention-based scanning strategy scans according to the attention weights of the instances in the sequence. (b) Illustration of fusion strategy in multimodal expert. Local token-level fusion and global distribution-level fusion are utilized to achieve efficient and effective multimodal fusion.}
    \label{figure1}
\end{figure*}

To address the above challenges, our objective is to leverage both the unimodal features of pathological images and genomic data, along with their interacted multimodal representations. This enables effective capture of critical information inherent in each individual modality while facilitating comprehensive survival analysis through integrated multimodal features. Moreover, to reduce computational complexity, we explore a more efficient sequence modeling framework—Mamba, which maintains strong modeling capabilities with linear computational complexity. By integrating a selection mechanism and hardware-aware parallel algorithms into structured state space models (SSMs), Mamba effectively captures long-range dependencies without the computational burden of attention mechanisms. Mamba has already been widely adopted in tasks such as WSIs classification\cite{20,21,22} and multimodal fusion\cite{23,24,25,26,27}. However, existing Mamba-based multiple instance learning methods often rely on multiple scanning directions to capture contextual relationships among instances, which may not adequately identify the most discriminative instance-level features, as illustrated in Fig. \ref{figure1} (a). Furthermore, current Mamba-based multimodal fusion approaches often simply interleave features from different modalities in sequence, lacking deeper interaction mechanisms that can comprehensively capture cross-modal relationships. This limitation ultimately compromises the quality of the learned cross-modal representations.

In this paper, we propose a multi-expert system named Multi-Expert Mamba (ME-Mamba), which integrates pathology images and genomics data for survival analysis. This system processes both unimodal data including pathology and genomics and their multimodal fusion in parallel. Such a parallel architecture facilitates an in-depth understanding of the potential factors influencing survival outcomes within both unimodal and multimodal representations. It achieves complementary integration of modalities without losing critical unimodal information, while the Mamba-based structure significantly improves computational efficiency. More specifically, the system consists of two unimodal experts (a Pathology Expert and a Genomics Expert), along with a Synergistic Expert for multimodal fusion. The unimodal experts are designed as attention-based mamba multiple instance learning modules. Each expert employs three scanning strategies. Specifically, an attention model first scores each instance in the sequence, which is then reordered by descending attention scores to explicitly capture discriminative key instances. The two conventional scanning directions serve as supplements to capture global contextual relationships among instances, as illustrated in Fig. \ref{figure1} (a). The Synergistic Expert is responsible for cross-modal interaction. We first use Optimal Transport to explicitly learn token-level local correspondences between the two modalities. A global cross-modal fusion loss based on Maximum Mean Discrepancy is applied to implicitly enhance distribution consistency. The fused representations are then forwarded to a mamba backbone for further integration. This multi-stage fusion strategy ensures that the model learns comprehensive multimodal representations, as shown in Fig. \ref{figure1} (b). Through the collaboration of the Pathology Expert, Genomics Expert, and Synergistic Expert, our method effectively captures discriminative information within each modality while enabling thorough multimodal fusion.

The main contributions of this paper can be summarized as follows:
\begin{itemize}
\item We propose a Multi-Expert Mamba (ME-Mamba) system, which for the first time enables parallel processing of pathology images, genomics data, and their multimodal fusion, effectively overcoming limitations inherent in transformer-based architectures.
\item We introduce an attention-guided Mamba architecture (comprising a Pathology Expert and a Genomics Expert) for modeling pathological and genomic modalities. This design explicitly captures both discriminative key features and global contextual information within each unimodal representation.
\item We develop a multimodal mamba architecture (the Synergistic Expert) that incorporates both local token-level alignment and global distribution consistency, enabling comprehensive cross-modal interaction.
\item We evaluate ME-Mamba’s performance on five public TCGA datasets, including BLCA, BRCA, UCEC, GBMLGG, LUAD. ME-Mamba achieves superior performance in survival prediction task, notably outperforming all comparative methods by a large margin (8\% on average).
\end{itemize}

\section{Related work}
\subsection{State space models}
% \subsubsection{Traditional Image Fusion Algorithms}
State Space Models (SSMs) are a class of sequence modeling frameworks that capture temporal or sequential dependencies through latent state transitions, making them particularly suitable for processing long sequences. Unlike the self-attention mechanism in Transformers, which has quadratic computational complexity, SSMs model sequences via a series of latent state updates, achieving linear complexity. S4\cite{28} addressed the high computational and memory costs of earlier SSMs by stabilizing the diagonalization of the state matrix and simplifying computation via the Cauchy kernel. S5\cite{29} further improved efficiency by replacing multiple independent single-input single-output (SISO) SSMs with a multi-input multi-output (MIMO) SSM, eliminating the need for complex convolutional kernel computation. Mamba\cite{30} (S6) enhanced S4 by incorporating an input-dependent selection mechanism and hardware-aware algorithms. Compared to Transformers with quadratic-complexity attention, Mamba excels at processing long sequences with linear complexity.

Leveraging the advancements of SSMs, research interest in their application to computer vision tasks has surged. Vim\cite{31} first introduced Mamba to computer vision tasks and proposed a bidirectional scanning strategy to capture contextual dependencies. However, such a scanning strategy fails to capture extensive spatial interactions owing to its constrained scanning directions. Later, VMamba\cite{32} designed a 2D selective scanning (SS2D) module that traverses image patches along four scanning paths to enlarge the receptive field. EfficientVMamba\cite{33} further introduced a lightweight interleaved scanning mechanism to approximate SS2D, thereby reducing computational cost.

In the context of WSIs analysis, Mamba-based methods have also demonstrated significant advantages. MambaMIL\cite{20} first integrated the Mamba framework with multiple instance learning for WSI analysis, enhancing spatial interaction by rearranging sequences from top to bottom in addition to the original scanning direction. To further improve spatial interaction, MSMMIL\cite{22} introduced grid scanning and layer scanning strategies, enabling four-directional modeling within a single sequence length and achieving efficient long-sequence modeling. 2DMamba\cite{21} designed a novel 2D scanning approach to directly process 2D feature maps, better preserving the spatial continuity of images. However, all these methods primarily focus on capturing global sequence information and improving spatial interaction, without explicitly emphasizing discriminative key information within long sequences. In contrast, our approach introduces an attention-guided scanning mechanism that explicitly prioritizes discriminative information by scanning tokens in descending order of their attention scores. Combined with conventional left-to-right and top-to-bottom scanning directions, our method effectively captures both global contextual relationships and discriminative features from long instance sequences with high efficiency.

\subsection{Survival analysis}

\paragraph{Unimodal}Survival analysis is a statistical and modeling method used to assess the time of event occurrence and related influencing factors, which can provide valuable information for doctors to evaluate clinical outcomes of disease progression and treatment efficacy. Traditional survival analysis methods are predominantly unimodal. In early research, survival analysis primarily relied on Cox proportional hazards models\cite{34}, utilizing short-term clinical indicators and long-term follow-up reports\cite{4,5,6,7,35}. For example, Lai et al.\cite{5} combined systems biology and deep learning with fifteen biomarkers and clinical data to predict survival outcomes in non-small cell lung cancer (NSCLC) patients. Yu et al.\cite{35} integrated baseline prognostic parameters with dynamic trends in clinical indicators to develop a dynamic survival prediction model for acute-on-chronic liver failure (ACLF) patients. Pathology images directly reveal morphological characteristics of tumors and provide valuable information regarding tumor aggressiveness, treatment response, and likelihood of disease recurrence. Early studies focusing on pathology image analysis mainly addressed tasks such as classification\cite{36,37,38,39,40,41}, though several works also utilized pathological images for survival analysis\cite{8,9,10,11,12,42}. For instance, WSISA\cite{8} achieved end-to-end survival prediction using pathological images via clustering and deep convolutional networks. DeepAttnMISL\cite{10} employed an attention-based multiple instance learning (MIL) mechanism to aggregate patient-level representations, effectively performing survival analysis on datasets. However, survival analysis models based solely on unimodal pathological images remain insufficient for clinical applications. With the rapid advancement of high-throughput sequencing technologies, genomics data has demonstrated high relevance as a metric for disease modeling and prognosis\cite{43,44,45,46}, thereby opening a new avenue for survival analysis.

\paragraph{Multimodal}In recent years, multimodal data fusion has garnered increasing attention in survival analysis tasks. Multimodal data provides multidimensional insights into a patient’s condition at macroscopic, microscopic, and molecular levels. By integrating information from different modalities, a more comprehensive understanding of the disease can be achieved, leading to more accurate diagnoses, optimized treatment strategies, and more reliable prognostic predictions. The integration of pathology images and genomics data\cite{13,14,15,17,18,19,47,48, 49} enables the simultaneous capture of morphological characteristics and molecular profiles of tumors, thereby enhancing the predictive power of survival models. For instance, MCAT\cite{19} proposed an interpretable, dense co-attention mapping between WSIs and genomic features formulated in the embedding space. CMTA\cite{14} employed two parallel encoder-decoder structures to process pathological and genomic data separately, along with a cross-modal attention module serving as a bridge to explore inter-modal relationships and transfer complementary information. CCL\cite{13} designed four parallel Transformer encoders to explicitly decompose knowledge into four components for more effective multimodal integration. Existing methods generally rely on cross-attention or self-attention mechanisms to achieve cross-modal interaction and fusion, aiming to learn comprehensive and effective multimodal representations. However, high-dimensional multimodal features often obscure critical survival-related information present in the original unimodal data. Moreover, the quadratic time complexity of Transformers limits their efficiency when handling large-scale or long-sequence data. These limitations underscore the need to develop novel multimodal fusion approaches that balance effectiveness and efficiency. To address these challenges, we designed a Mamba-based multi-expert system that processes unimodal features in parallel and obtains multimodal representations through both local token-level fusion and global distribution alignment, while significantly improving computational efficiency.

\section{Methodology}
In this section, we first describe the preliminary knowledge of multiple instance learning (MIL) and state space models (SSMs), and then provide an overview of the proposed Multi-Expert Mamba system along with its core components.
\label{sec103}
\subsection{Preliminaries}
\subsubsection{MIL formulation}
Multiple Instance Learning (MIL) is a weakly supervised learning approach that models training data as bags, where each bag contains multiple instances. A bag is labeled as positive if it contains at least one positive instance. Conversely, a bag is negative only if all its instances are negative. For example, let $X_i=\left\{\left(x_{i, 1}, y_{i, 1}\right),\left(x_{i, 2}, y_{i, 2}\right), \ldots,\left(x_{i, j}, y_{i, j}\right), \ldots,\left(x_{i, n}, y_{i, n}\right)\right\}$ denote the $i$-th bag, where $n$ is the number of instances. The label of the bag is then defined as,
\begin{equation}
Y_i= \begin{cases}0, & \text { if } \sum_j y_{i, j}=0 \\ 1, & \text { else }\end{cases}
\end{equation}

In a typical MIL pipeline, instance-level features are first extracted. These features are then aggregated into bag-level representations using a specific pooling or attention mechanism. Finally, a bag-level classifier is applied to predict the label of the bag based on the aggregated features.

\subsubsection{State space models}
Inspired by SSMs, structured state space sequence (S4) models have emerged as an effective architecture for long sequence modeling. As a linear time-invariant system, the S4 model is parameterized by four components $(\Delta, A, B, C)$, mapping a one-dimensional input sequence $x(t) \in \mathbb{R}^L$ to an output $y(t) \in \mathbb{R}^L$ through a hidden state $h(t) \in \mathbb{R}^N$. This process can be described by the following continuous system,
\begin{equation}
h^{\prime}(t)=A h(t)+B x(t), y(t)=C h(t)
\end{equation}

where $A \in \mathbb{R}^{N \times N}$ is the state transition matrix. $B \in \mathbb{R}^{N \times 1}$ and $C \in \mathbb{R}^{N \times 1}$ are projection parameters. The S4 model uses a timescale parameter $\Delta$ to convert the continuous parameters $A$, $B$ into discrete counterparts $\bar{A}$, $\bar{B}$ via,
\begin{equation}
\bar{A}=\exp (\triangle A), \bar{B}=(\triangle A)^{-1}(\exp (\triangle A)-I) \cdot \triangle B
\end{equation}

After discretization, the model can be computed recurrently for efficient auto-regressive inference,
\begin{equation}
h_t=\bar{A} h_{t-1}+\bar{B} x_t, y_t=C h_t
\end{equation}

In practical training scenarios, the model can also be computed efficiently in parallel using a convolutional approach,
\begin{equation}
\bar{K}=\left(C \bar{B}, C \overline{A B}, \ldots, C \bar{A}^{M-1} \bar{B}\right), y=x * \bar{K}
\end{equation}

Mamba further enhances the S4 model by incorporating a selection mechanism and a hardware-aware parallel algorithm. This allows the model to overcome the limitations of static parameters in S4, enabling efficient long-sequence modeling through input-dependent selective propagation or forgetting of information along the sequence.

\subsection{Overview and data processing}
\subsubsection{Overview of the framework}
\begin{figure*}[t!]
    \centering
    \includegraphics[width=\textwidth]{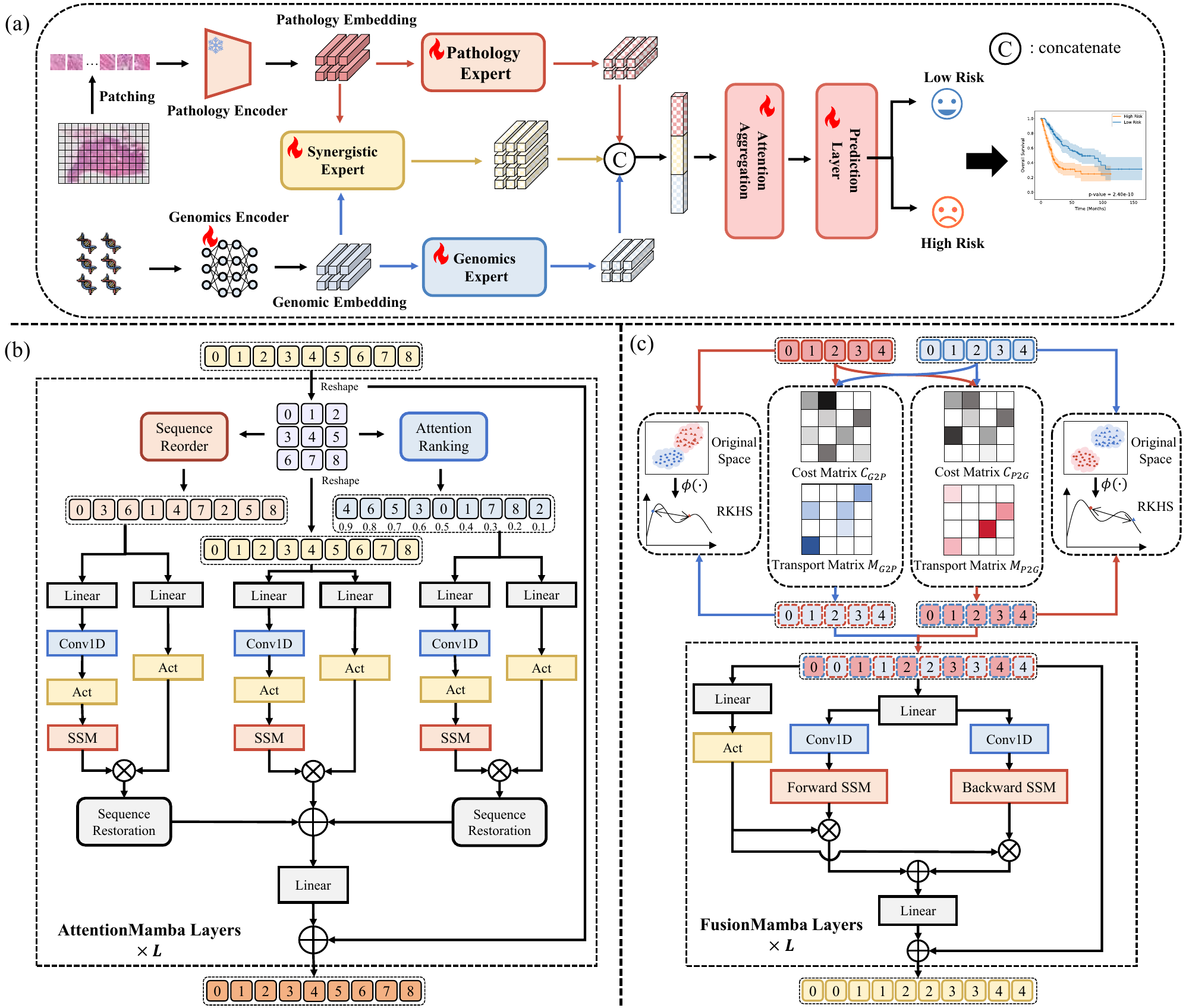}
    \caption{(a) An overview of the proposed Multi-Expert Mamba (ME-Mamba) system, which is composed of Pathology Expert, Genomics Expert and Synergistic Expert. (b) Schematic of the pathology and genomics expert architecture, which employs three concurrent scanning methods to capture both global context and discriminative features from long sequences. (c) Architecture of the Synergistic Expert, comprising feature fusion, feature scanning, and feature encoding stages.}
    \label{figure2}
\end{figure*}

The proposed Multi-Expert Mamba (ME-Mamba) system, as illustrated in Fig. \ref{figure2} (a), consists of three major steps: instance-level feature extraction, expert-based feature processing, and outcome prediction. In the first step, each pathological whole-slide image (WSI) is divided into thousands of non-overlapping patches, while genomic data are grouped by functional categories. Specific feature extraction methods are then applied to each modality, as detailed in Section 3.2.2. In the second step, a Pathology Expert and a Genomics Expert are utilized to extract discriminative features from the highly redundant pathological images and genomic data, respectively. A Synergistic Expert is further employed to effectively integrate the pathological and genomic features. These modules are elaborated in Sections 3.3 and 3.4. Finally, the extracted unimodal features and the fused multimodal features are combined for final survival prediction, as described in Section 3.5.

\subsubsection{Feature extraction}
\paragraph{Pathology}Pathology images, namely Whole Slide Images (WSIs), provide morphological information about the tumor immune microenvironment. However, due to their extremely high resolution, WSIs cannot be directly processed by convolutional neural networks and must first be partitioned. We begin by segmenting the tissue regions, then extract non-overlapping patches of size 256×256 at 20× magnification. Following previous works\cite{13,14,18}, we employ a pre-trained ResNet50\cite{50} model, which was initially trained on ImageNet, to extract a 1024-dimensional embedding for each patch. All patch embeddings of the same WSI are collected as an embedding set. To reduce feature redundancy and computational overhead, we employ a Multi-Layer Perceptron (MLP) to reduce the feature dimension from 1024 to 256. The resulting feature vectors are then passed to the Pathology Expert Mamba module for feature aggregation.

\paragraph{Genomics}Genomics profiles can identify specific genetic alterations or biomarkers associated with cancer prognosis. Certain gene mutations, gene expression patterns, and DNA copy number variations may serve as prognostic indicators, aiding in the prediction of patient survival outcomes. We partition RNA sequencing (RNA-seq), Copy Number Variation (CNV), and Simple Nucleotide Variation (SNV) sequences into six sub-sequences as previous methods\cite{13,14,18}]. Each subsequence is transformed into a feature vector using a two-layer Self-normalizing Neural Network (SNN)\cite{51}, and a Multi-Layer Perceptron (MLP) is then applied to obtain a 256-dimensional representation. These feature vectors are subsequently passed to the Genomics Expert Mamba module for feature aggregation.

\subsection{Pathology expert and genomics expert}
The Pathology Expert is designed to aggregate instance-level features from pathological images. It comprises multiple stacked Attention-based Mamba Layers, which extract discriminative features and capture global information from long instance sequences containing substantial redundant or irrelevant data, as illustrated in Fig. \ref{figure2} (b). By integrating the Mamba structure with multiple instance learning, each instance can interact with previously scanned instances via compressed hidden states, enabling effective modeling of long sequences while reducing computational complexity. We combine conventional scanning strategies with an attention-guided scanning mechanism, forming a three-branch parallel Mamba architecture. The conventional scanning includes two strategies: the original scan and the transposed scan. The attention-based scanning mechanism ranks the instances according to their attention scores, allowing the model to explicitly focus on the most discriminative features.

In detail, given instance features $X \in \mathbb{R}^{n \times d}$, where $n$ is the sequence length and $d$ is the feature dimension, we first feed the sequence into three parallel branches. In the first branch, the original sequence order is preserved and passed to subsequent Casual Convolution and State Space Model (SSM) layers for sequence modeling. In the second branch, following MambaMIL\cite{20}, the instance sequence is transposed and scanned sequentially before being processed by the same network layers. In the third branch, we use the attention mechanism from ABMIL\cite{52} to assign an attention score to each instance. The instances are then reordered in descending order of their attention scores, allowing the model to prioritize highly-attended features in subsequent processing. The overall procedure is formulated as follows,
\begin{equation}
X_r=S R(X), X_a=A R(X)
\end{equation}
\begin{equation}
X^{\prime}=\operatorname{Norm}(X), X_r^{\prime}=\operatorname{Norm}\left(X_r\right), X_a^{\prime}=\operatorname{Norm}\left(X_a\right)
\end{equation}
\begin{equation}
Y=\operatorname{SSM}\left(\operatorname{SiLU}\left(\operatorname{Conv} 1 D\left(\operatorname{Linear}\left(X^{\prime}\right)\right)\right)\right.
\end{equation}
\begin{equation}
Y_r=\operatorname{SSM}\left(\operatorname{SiLU}\left(\operatorname{Conv} 1 D\left(\operatorname{Linear}\left(X_r^{\prime}\right)\right)\right)\right.
\end{equation}
\begin{equation}
Y_a=\operatorname{SSM}\left(\operatorname{SiLU}\left(\operatorname{Conv} 1 D\left(\operatorname{Linear}\left(X_a^{\prime}\right)\right)\right)\right.
\end{equation}
\begin{equation}
Z=\operatorname{SiL} U\left(\operatorname{Linear}\left(X^{\prime}\right)\right), Z_r=\operatorname{SiLU}\left(\operatorname{Linear}\left(X_r^{\prime}\right)\right), Z_a=\operatorname{SiLU}\left(\operatorname{Linear}\left(X_a^{\prime}\right)\right)
\end{equation}
\begin{equation}
X^{\prime \prime}=Z \odot Y, Y_r^{\prime}=Z_r \odot Y_r, Y_a^{\prime}=Z_a \odot Y_a
\end{equation}
\begin{equation}
X_r^{\prime \prime}=\varphi\left(Y_r^{\prime}\right), X_a^{\prime \prime}=\psi\left(Y_a^{\prime}\right)
\end{equation}
\begin{equation}
X_{\text {output }}=\operatorname{Linear}\left(X^{\prime \prime}+X_r^{\prime \prime}+X_a^{\prime \prime}\right)+X
\end{equation}

where $S R(\cdot)$ denotes the transposed scan ordering, and $\varphi(\cdot)$ represents sequence restoration. $A R(\cdot)$ denotes reordering by attention score, and $\psi(\cdot)$ indicates sequence restoration.

The Genomics Expert follows the same computational process as the Pathology Expert but uses separate parameters. Through this approach, the model preserves the original sequence order and distribution, reconstructs features from a global perspective, and emphasizes discriminative instances via attention-based reordering. This enables the simultaneous capture of both global contextual information and key instance-level features.

\subsection{Synergistic Expert}
The Synergistic Expert is designed to effectively integrate features from both pathological images and genomic data. It incorporates two complementary fusion mechanisms: a local token-level alignment based on Optimal Transport (OT)\cite{65}, and a global distribution matching via Maximum Mean Discrepancy (MMD)\cite{66}. These are applied bidirectionally, by using both pathology and genomics features as anchors. The resulting cross-modality-aware features are then passed through multiple stacked Multimodal Mamba Layers to further enhance fusion, as illustrated in Fig. \ref{figure2} (c).
\subsubsection{Local cross-modal fusion}
We employ Optimal Transport to achieve fine-grained, token-level alignment between the two modalities, using each in turn as an anchor. This approach treats the feature sequences as discrete distributions and learns a transport plan that minimizes the cost of mapping one distribution to the other, thereby establishing token-wise correspondences.

In detail, given pathology instance features $X_p \in \mathbb{R}^{n \times d}$  and genomics instance features $X_g \in \mathbb{R}^{m \times d}$, where $n$ and $m$ are the sequence lengths and $d$ is the feature dimension, our goal is to learn a transport matrix $M$ that captures fine-grained inter-modal relationships. Taking the mapping from pathology to genomics (with genomics as anchor) as an example, the objective is formulated as,
\begin{equation}
\min _{p 2 g} \sum_{i=1}^n \sum_{j=1}^m M_{p 2 g}(i, j) C_{p 2 g}(i, j)
\end{equation}

where $C_{p 2 g} \in \mathbb{R}^{n \times m}$ is the cost matrix. We use cosine distance to emphasize angular similarity between feature vectors, which also offers numerical stability due to its bounded range,
\begin{equation}
C_{p 2 g}(i, j)=1-\frac{X_p^i \cdot X_g^j}{\left\|X_p^i\right\|_2\left\|X_g^j\right\|_2}
\end{equation}

As solving this optimal transport problem can be computationally expensive, we adopt a simplified version following prior methods\cite{53,54}, 
\begin{equation}
\left\{\begin{array}{c}
\sum_{j=1}^m M_{p 2 g}(i, j)=\frac{1}{n}, \forall i \in[1, n] \\
M_{p 2 g}(i, j) \geq 0, \forall i, j
\end{array}\right.
\end{equation}

This simplification allows one genomic instance to interact with multiple pathological instances, maintaining the ability to capture meaningful cross-modal correlations while significantly reducing computational complexity. The transport matrix is ultimately computed as,
\begin{equation}
M_{p 2 g}(i, j)=\left\{\begin{array}{l}
\frac{1}{n}, j=\operatorname{argmin}_j C_{p 2 g}(i, j) \\
0, j \neq \operatorname{argmin}_j C_{p 2 g}(i, j)
\end{array}\right.
\end{equation}

Similarly, we compute the transport matrix $M_{g 2 p}$ for the genomics-to-pathology mapping. The fused features are then obtained as,
\begin{equation}
\left\{\begin{array}{l}
X_p^{\prime}=M_{p 2 g}^T X_p \in \mathbb{R}^{m \times d} \\
X_g^{\prime}=M_{g 2 p}^T X_g \in \mathbb{R}^{n \times d}
\end{array}\right.
\end{equation}

While this OT-based fusion effectively captures local token-level correspondences, it does not inherently ensure global consistency between the modalities. To address this, we introduce an additional global cross-modal fusion strategy.

\subsubsection{Global cross-modal fusion}
To implicitly align the global distributions between locally fused features and anchor features, we employ Maximum Mean Discrepancy (MMD). MMD measures the statistical divergence between different modalities by comparing their statistics in a high-dimensional Reproducing Kernel Hilbert Space (RKHS). For two feature $X$ and $Y$, the squared MMD distance is defined as,
\begin{equation}
M M D^2(X, Y)=\left\|\frac{1}{N} \sum_{i=1}^N \phi\left(x_i\right)-\frac{1}{N} \sum_{j=1}^N \phi\left(y_i\right)\right\|_{\mathcal{H}}^2
\end{equation}

where $\phi(\cdot)$ denotes the mapping from the original feature space to the RKHS $\mathcal{H}$. In practice, we compute this using a kernel function,
\begin{equation}
M M D^2(X, Y)=\frac{1}{N^2} \sum_{i=1}^N \sum_{i^{\prime}=1}^N k\left(x_i, x_{i^{\prime}}\right)+\frac{1}{N^2} \sum_{j=1}^N \sum_{j^{\prime}=1}^N k\left(y_i, y_{i^{\prime}}\right)-\frac{2}{N^2} \sum_{i=1}^N \sum_{j=1}^N k\left(x_i, y_i\right)
\end{equation}

Specifically, we use a Gaussian kernel $k(x, y)=\exp \left(-\frac{\|x-y\|_2^2}{2 \sigma^2}\right)$.

For the locally fused pathological features $X_p^{\prime}$ $X_g^{\prime}$, the global alignment loss is formulated as,
\begin{equation}
L_{\text {global }}=M M D^2\left(X_p^{\prime}, X_g\right)+M M D^2\left(X_g^{\prime}, X_p\right)
\end{equation}

Minimizing this loss during training ensures global distributional consistency between the modalities after fusion. Our strategy, which combines explicit local token-level alignment with implicit global distribution matching, enables multi-granular alignment and facilitates more effective feature encoding in the subsequent Mamba backbone.

\subsubsection{ Multimodal Mamba fusion}
After performing token-level explicit local fusion and distribution-level implicit global alignment between the pathological and genomic instance features, we employ a bidirectional Mamba (BiMamba)\cite{31} backbone to further integrate the two modalities. Unlike traditional Transformer-based methods that process all tokens simultaneously via self-attention, our approach adopts an ordered scanning strategy that preserves the sequential nature of Mamba while enabling effective cross-modal interaction. Given the locally and globally fused feature sequences, we construct a unified multimodal feature sequence by interleaving features from the two modalities in order,
\begin{equation}
X_{\text {fusion }}=\left[X_{p_1}^{\prime}, X_{g 1}^{\prime}, X_{p_2}^{\prime}, X_{g 2}^{\prime}, \ldots, X_{p m}^{\prime}, X_{g m}^{\prime}, \ldots, X_{p n}^{\prime}\right]
\end{equation}

This interleaved organization ensures that features from different modalities are processed sequentially, allowing Mamba’s selective scanning mechanism to effectively capture both intra-modal and inter-modal dependencies. The multimodal representation is ultimately obtained through multiple stacked BiMamba layers.

\subsection{Feature aggregation and prediction}
After processing by the respective expert modules, the pathological instance feature sequence, genomic feature sequence, and multimodal feature sequence, each of which contains discriminative information, are obtained. We concatenate these three feature sequences and aggregate the instance-level features into a bag-level representation. The aggregation method follows ABMIL\cite{52}. Finally, a Multi-Layer Perceptron (MLP) is used to predict the hazard function $H$,
\begin{equation}
\left.H=M L P\left(A G G\left(C A T\left(C A T\left(X_p, X_g\right), X_{\text {fusion }}\right)\right)\right)\right)
\end{equation}

where $X_p$, $X_g$ and $X_fusion$ enote the feature sequences processed by the Pathology Expert, Genomics Expert, and Synergistic Expert, respectively. $CAT$ represents the concatenation operation, andAGG denotes feature aggregation.

For survival prediction, following previous works\cite{13,14,18,19,47,48}, we simplify the original event time regression problem to a classification problem. The ground truth survival time of a patient is discretized into $n$ equal intervals. The interval $t_k$ in which the event occurs is used as the class label $k$ for the patient. The model predicts the probability of the event occurring in each time interval, forming a hazard vector hazard vector $H=\left\{h_1, \ldots, h_k, \ldots, h_n\right\}$.Each patient sample is represented as a triple $\{H, c, k\}$, where $c \in\{0,1\}$ indicates the right uncensorship status. The discrete survival function is defined as $f_{\text {surv }}(H, k)=\prod_{i=1}^k\left(1-h_i\right)$. The survival prediction loss is formulated as,
\begin{equation}
L_{\text {surv }}=-c \log \left(f_{\text {surv }}(H, k)\right)-(1-c) \log \left(f_{\text {surv }}(H, k-1)\right)-(1-c) \log \left(h_k\right)
\end{equation}

Finally, the overall loss function of our framework is formulated as,
\begin{equation}
L=L_{\text {surv }}+\lambda L_{\text {global }}
\end{equation}

where $\lambda$ is a weighting coefficient for the global alignment loss.

\section{Experiments and results}
\label{sec104}
In this section, we conduct extensive experiments on five public datasets to evaluate the effectiveness of our proposed model. We first introduce the datasets and evaluation metrics used in the study. Then, we compare the experimental results with several state-of-the-art methods to demonstrate the superiority of our model, along with an interpretability analysis. Finally, we perform ablative experiments to investigate the impact of key components.

\subsection{Experimental settings}
\paragraph{Datasets}To verify the performance of the proposed method, we carried out a series of experiments using five cancer datasets. These datasets are derived from The Cancer Genome Atlas (TCGA)\footnote{https://portal.gdc.cancer.gov/} that contains paired diagnostic Whole Slide Images (WSIs) and genomics data, along with clinical information from thousands of cancer patients. These include Bladder Urothelial Carcinoma (BLCA, n = 372), Breast Invasive Carcinoma (BRCA, n = 956), Uterine Corpus Endometrial Carcinoma (UCEC, n = 480), Glioblastoma \& Lower Grade Glioma (GBMLGG, n = 569), and Lung Adenocarcinoma (LUAD, n = 453). For WSIs, we first segment the tissue regions of each slide and then cut them into 256×256 patches at 20× magnification. For genomic data, following previous works[13], [14], [19], we use RNA-seq, CNV, and SNV sequences and further group them into six subsequences: 1) Tumor Suppression, 2) Oncogenesis, 3) Protein Kinases, 4) Cellular Differentiation, 5) Transcription, and 6) Cytokines and Growth.
\paragraph{Evaluation Metrics}The concordance index (C-index)\cite{55} is adopted to measure the survival prediction performance. C-index measures a model’s ability to accurately rank individuals’ survival times in survival analysis, assessing the concordance between predicted risk scores and actual survival outcomes. C-index can be formulated as follows,
\begin{equation}
c-\text { index }=\frac{1}{n(n-1)} \sum_{i=1}^n \sum_{j=1}^n I\left(T_i<T_j\right)\left(1-c_j\right)
\end{equation}

where $n$ is the number of patients, $T_i$ and $T_j$ are the survival times of $i$-th and $j$-th patient. $I(\cdot)$ denotes the indicator function, which evaluates to 1 if the enclosed condition is true and to 0 otherwise. 

\paragraph{Implementation Details}We employed 5-fold cross-validation to evaluate our model and other compared methods in five cancer survival prediction tasks. Specifically, we first randomly shuffle the dataset and split it into five groups, with four serving as training sets and one as a test set. We train the models on the training set and evaluate their performance on the test set to report the corresponding C-index scores of mean ± std (standard deviation). We adopted the SGD optimizer with a learning rate of 1e-3, and the framework was trained for 30 epochs. All experiments were conducted using Python 3.10 with Pytorch toolkit version 2.0 on a platform equipped with NVIDIA GeForce RTX 4090 GPUs.

\subsection{Comparisons with state-of-the-art approaches}
To demonstrate the effectiveness of ME-Mamba, we compare it with unimodal baselines and multimodal SOTA methods. For genomics data, we implemented SNN\cite{51} and SNNTrans. For pathology data, we implemented ABMIL\cite{52}, CLAM\cite{56}, TransMIL\cite{57} and DTFD\cite{58}. For multimodal models, we chose MCAT\cite{19}, M3IF\cite{59}, GPDBN\cite{15}, Porpoise\cite{60}, HFBSurv\cite{61}, SurvPath\cite{48}, MOTCat\cite{18}, CMTA\cite{14} and CCL\cite{13}. These models are categorized into unimodal and multimodal groups, and several representative ones are briefly introduced below.

\paragraph{Unimodal Model}For genomics data, SNNTrans is a variant of SNN\cite{51} that uses a Self-normalizing Neural Network (SNN) to extract instance-level genomic features, followed by TransMIL to aggregate them into bag-level representations. For pathology data, ABMIL assumes that image instances are independent and identically distributed (i.i.d.) and employs an attention mechanism to aggregate instance features. TransMIL breaks the i.i.d. assumption by incorporating correlation modeling and spatial encoding, using self-attention to aggregate instance features.
\paragraph{Multimodal Model}MCAT uses a co-attention mechanism to dynamically align pathological image features and genomic features in an embedding space, followed by a Transformer for multimodal fusion. The aggregated features are then concatenated for survival prediction. SurvPath decomposes transcriptomic data into different biological pathway signatures and uses a sparse-attention Transformer to model interactions between pathways, pathological images, and across pathways. MOTCat employs optimal transport to compute a global matching flow between pathological image features and genomic features, capturing spatial interactions in the tumor microenvironment and structural consistency in gene co-expression more effectively than traditional co-attention. CCL explicitly decomposes interactive knowledge from pathological and genomic data and employs cohort-guided supervision at both the knowledge level and patient level.
\paragraph{Comparison with single-modal models}As shown in Table \ref{table1}, the proposed method achieves superior performance across all five datasets. Specifically, it attains a C-index of 0.6993 on BLCA, outperforming the best unimodal model by 8.3\%; 0.6910 on BRCA, showing a 6.7\% improvement; 0.7063 on UCEC, with a 2.4\% gain; 0.8669 on GBMLGG, exceeding the best unimodal result by 3.6\%; and 0.7014 on LUAD, representing a 10.7\% improvement. These results indicate that our method effectively integrates multimodal data and underscores the benefit of multimodal learning for survival analysis. Additionally, we observe that unimodal methods using genomic data generally outperform those using pathological images, suggesting that genomic features may exhibit stronger correlation with patient survival outcomes. Our approach successfully leverages complementary information from both modalities, further enhancing the accuracy of survival prediction.

\begin{table}[]
\centering
\caption{Survival analysis on five TCGA datasets. Bolded red and bolded blue are used to denote the best and the second best values, respectively.}
\resizebox{\textwidth}{!}{%
\begin{tabular}{ccccccccc}
\hline
Model         & G.                   & P.                   & BLCA                                          & BRCA                                          & UCEC                                          & GBMLGG                                        & LUAD                                          & Overall                                \\ \hline
\multicolumn{9}{c}{\textbf{Pathology feature   extraction backbone -- ResNet-50}}                                                                                                                                                                                                                                                                    \\ \hline
SNN           & \checkmark                    & \multicolumn{1}{l}{} & 0.6339±0.0509                                 & 0.6327±0.0739                                 & 0.6900±0.0389                                 & 0.8370±0.0276                                 & 0.6171±0.0411                                 & 0.6821                                 \\
SNNTrans      & \checkmark                    & \multicolumn{1}{l}{} & 0.6456±0.0428                                 & 0.6478±0.0580                                 & 0.6324±0.0324                                 & 0.8284±0.0158                                 & 0.6335±0.0493                                 & 0.6775                                 \\ \hline
ABMIL         & \multicolumn{1}{l}{} & \checkmark                    & 0.5673±0.0498                                 & 0.5899±0.0472                                 & 0.6507±0.0330                                 & 0.7974±0.0336                                 & 0.5753±0.0744                                 & 0.6361                                 \\
CLAM-SB       & \multicolumn{1}{l}{} & \checkmark                    & 0.5487±0.0286                                 & 0.6091±0.0329                                 & 0.6780±0.0342                                 & 0.7969±0.0346                                 & 0.5962±0.0558                                 & 0.6458                                 \\
CLAM-MB       & \multicolumn{1}{l}{} & \checkmark                    & 0.5620±0.0313                                 & 0.6203±0.0520                                 & 0.6821±0.0646                                 & 0.7986±0.0320                                 & 0.5918±0.0591                                 & 0.6510                                 \\
TransMIL      & \multicolumn{1}{l}{} & \checkmark                    & 0.5466±0.0334                                 & 0.6430±0.0368                                 & 0.6799±0.0304                                 & 0.7916±0.0272                                 & 0.5788±0.0303                                 & 0.6480                                 \\
DTFD          & \multicolumn{1}{l}{} & \checkmark                    & 0.5662±0.0353                                 & 0.5975±0.0406                                 & 0.6308±0.0190                                 & 0.7641±0.0297                                 & 0.5580±0.0404                                 & 0.6233                                 \\ \hline
MCAT          & \checkmark                    & \checkmark                    & 0.6727±0.0320                                 & 0.6590±0.0418                                 & 0.6336±0.0506                                 & 0.8350±0.0233                                 & 0.6597±0.0279                                 & 0.692                                  \\
M3IF          & \checkmark                    & \checkmark                    & 0.6361±0.0197                                 & 0.6197±0.0707                                 & 0.6672±0.0293                                 & 0.8238±0.0170                                 & 0.6299±0.0312                                 & 0.6753                                 \\
GPDBN         & \checkmark                    & \checkmark                    & 0.6354±0.0252                                 & 0.6549±0.0332                                 & 0.6839±0.0529                                 & 0.8510±0.0243                                 & 0.6400±0.0478                                 & 0.6930                                 \\
Porpoise      & \checkmark                    & \checkmark                    & 0.6461±0.0338                                 & 0.6207±0.0544                                 & 0.6918±0.0488                                 & 0.8479±0.0128                                 & 0.6403±0.0412                                 & 0.6894                                 \\
HFBSurv       & \checkmark                    & \checkmark                    & 0.6398±0.0277                                 & 0.6473±0.0346                                 & 0.6421±0.0445                                 & 0.8383±0.0128                                 & 0.6501±0.0495                                 & 0.6835                                 \\
SurvPath      & \checkmark                    & \checkmark                    & 0.6581±0.0357                                 & 0.6306±0.0340                                 & 0.6636±0.0354                                 & 0.8422±0.0161                                 & 0.6600±0.0233                                 & 0.6909                                 \\
MOTCat        & \checkmark                    & \checkmark                    & 0.6830±0.0260                                 & 0.6730±0.0060                                 & 0.6750±0.0400                                 & 0.8490±0.0280                                 & 0.6700±0.0380                                 & 0.7100                                 \\
CMTA          & \checkmark                    & \checkmark                    & {\color[HTML]{0070C0} \textbf{0.6910±0.0426}} & 0.6679±0.0434                                 & 0.6975±0.0409                                 & 0.8531±0.0116                                 & 0.6864±0.0359                                 & 0.7192                                 \\
CCL           & \checkmark                    & \checkmark                    & 0.6862±0.0253                                 & {\color[HTML]{0070C0} \textbf{0.6840±0.0339}} & {\color[HTML]{0070C0} \textbf{0.7026±0.0475}} & {\color[HTML]{0070C0} \textbf{0.8614±0.0149}} & {\color[HTML]{0070C0} \textbf{0.6957±0.0231}} & {\color[HTML]{0070C0} \textbf{0.7260}} \\
\textbf{Ours} & \checkmark                    & \checkmark                    & {\color[HTML]{EE0000} \textbf{0.6993±0.0270}} & {\color[HTML]{EE0000} \textbf{0.6909±0.0378}} & {\color[HTML]{EE0000} \textbf{0.7063±0.0248}} & {\color[HTML]{EE0000} \textbf{0.8669±0.0147}} & {\color[HTML]{EE0000} \textbf{0.7014±0.0236}} & {\color[HTML]{EE0000} \textbf{0.7330}} \\ \hline
\multicolumn{9}{c}{\textbf{Pathology feature   extraction backbone -- UNI}}                                                                                                                                                                                                                                                                          \\ \hline
MCAT          & \checkmark                    & \checkmark                    & 0.6510±0.0291                                 & 0.6640±0.0425                                 & 0.7007±0.0556                                 & 0.8515±0.0289                                 & 0.6566±0.0306                                 & 0.7047                                 \\
SurvPath      & \checkmark                    & \checkmark                    & 0.6290±0.0376                                 & 0.7005±0.0521                                 & 0.7376±0.0475                                 & 0.8286±0.0204                                 & 0.6321±0.0550                                 & 0.7056                                 \\
MOTCat        & \checkmark                    & \checkmark                    & {\color[HTML]{0070C0} \textbf{0.6525±0.0280}} & 0.6651±0.0421                                 & 0.7383±0.0552                                 & 0.8491±0.0154                                 & 0.6648±0.0275                                 & 0.7147                                 \\
CMTA          & \checkmark                    & \checkmark                    & 0.6521±0.0318                                 & 0.7059±0.0304                                 & 0.7403±0.0432                                 & {\color[HTML]{0070C0} \textbf{0.8541±0.0207}} & 0.6642±0.0462                                 & 0.7233                                 \\
CCL           & \checkmark                    & \checkmark                    & 0.6404±0.0314                                 & {\color[HTML]{0070C0} \textbf{0.7252±0.0410}} & {\color[HTML]{EE0000} \textbf{0.7631±0.0482}} & 0.8536±0.0126                                 & {\color[HTML]{EE0000} \textbf{0.6737±0.0395}} & {\color[HTML]{0070C0} \textbf{0.7288}} \\
\textbf{Ours} & \checkmark                    & \checkmark                    & {\color[HTML]{EE0000} \textbf{0.6583±0.0231}} & {\color[HTML]{EE0000} \textbf{0.7313±0.0306}} & {\color[HTML]{0070C0} \textbf{0.7508±0.0393}} & {\color[HTML]{EE0000} \textbf{0.8624±0.0191}} & {\color[HTML]{0070C0} \textbf{0.6695±0.0490}} & {\color[HTML]{EE0000} \textbf{0.7345}} \\ \hline
\multicolumn{9}{c}{\textbf{Pathology feature   extraction backbone -- CONCH}}                                                                                                                                                                                                                                                                        \\ \hline
MCAT          & \checkmark                    & \checkmark                    & 0.6605±0.0159                                 & 0.6372±0.0571                                 & 0.7075±0.0373                                 & 0.8438±0.0117                                 & 0.6640±0.0275                                 & 0.7026                                 \\
SurvPath      & \checkmark                    & \checkmark                    & 0.6422±0.0194                                 & 0.6740±0.0445                                 & 0.7477±0.0333                                 & 0.8447±0.0237                                 & 0.6438±0.0493                                 & 0.7105                                 \\
MOTCat        & \checkmark                    & \checkmark                    & 0.6662±0.0327                                 & 0.6526±0.0294                                 & 0.7160±0.0319                                 & 0.8442±0.0188                                 & 0.6736±0.0371                                 & 0.7166                                 \\
CMTA          & \checkmark                    & \checkmark                    & 0.6694±0.0148                                 & {\color[HTML]{EE0000} \textbf{0.7273±0.0613}} & 0.7431±0.0484                                 & {\color[HTML]{0070C0} \textbf{0.8518±0.0206}} & 0.6750±0.0100                                 & 0.7333                                 \\
CCL           & \checkmark                    & \checkmark                    & {\color[HTML]{0070C0} \textbf{0.6701±0.0323}} & 0.6958±0.0431                                 & {\color[HTML]{EE0000} \textbf{0.7649±0.0343}} & 0.8454±0.0186                                 & {\color[HTML]{0070C0} \textbf{0.6868±0.0253}} & {\color[HTML]{0070C0} \textbf{0.7326}} \\
\textbf{Ours} & \checkmark                    & \checkmark                    & {\color[HTML]{EE0000} \textbf{0.6788±0.0265}} & {\color[HTML]{0070C0} \textbf{0.7166±0.0482}} & {\color[HTML]{0070C0} \textbf{0.7596±0.0438}} & {\color[HTML]{EE0000} \textbf{0.8593±0.0216}} & {\color[HTML]{EE0000} \textbf{0.6921±0.0350}} & {\color[HTML]{EE0000} \textbf{0.7413}} \\ \hline
\end{tabular}}
\label{table1}
\end{table}

\begin{figure*}[t!]
    \centering
    \includegraphics[width=\textwidth]{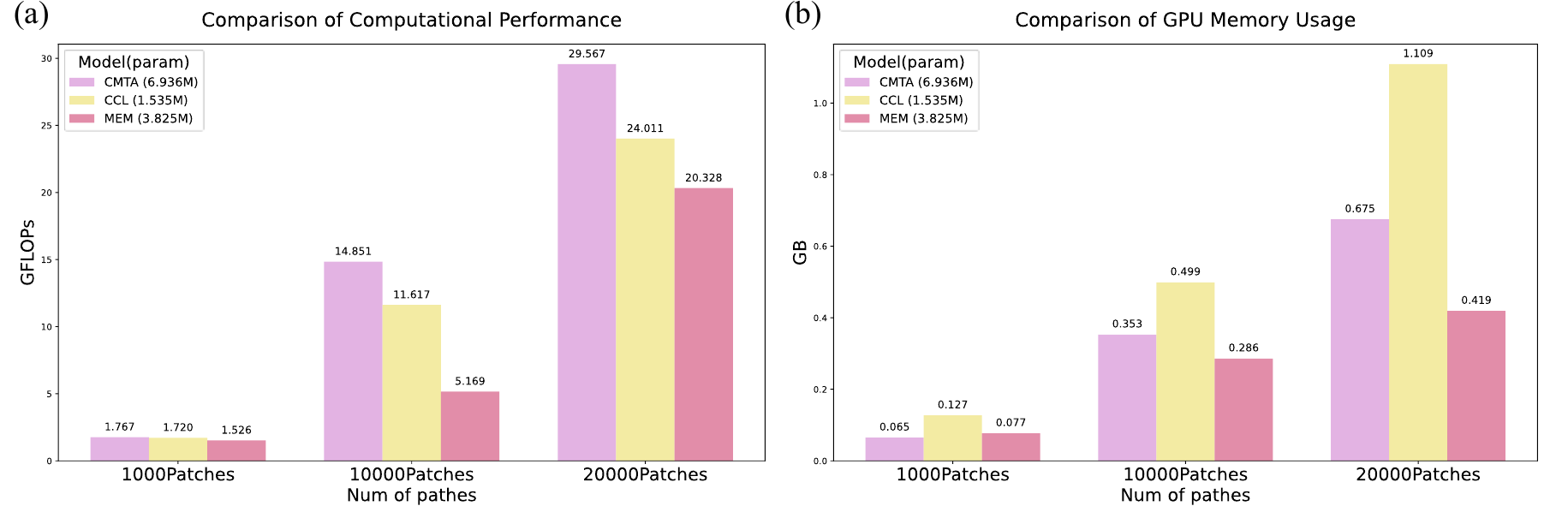}
    \caption{(a) Computational efficiency comparison with varying patches. (b) GPU memory usage comparison with varying patches.}
    \label{figure3}
\end{figure*}

\paragraph{Comparison with multi-modal models}As shown in Table \ref{table1}, our method also achieves superior performance compared to the best existing multimodal approaches. Specifically, it improves the C-index by 1.2\% on BLCA, 1.1\% on BRCA, 0.5\% on UCEC, 0.6\% on GBMLGG, and 0.8\% on LUAD over the strongest multimodal baseline. This improvement can be attributed to our multi-expert architecture, which explicitly captures discriminative information while preserving global context and enabling effective modality fusion. 

Furthermore, unlike existing Transformer-based multimodal methods, our approach leverages the Mamba architecture, leading to higher computational efficiency and reduced memory consumption. We compared ME-Mamba with two top-performing Transformer-based methods, CMTA and CCL, under varying numbers of instances, as illustrated in Fig. \ref{figure3}. We manually constructed instance vectors of dimension 1024 with counts of 1000, 10,000, and 20,000 to simulate different numbers of image patches extracted from WSIs, while keeping the number of genomic feature groups fixed at six. First, we compared GPU memory usage across different instance quantities (Fig. \ref{figure3} (a)). ME-Mamba consistently consumed the least memory across all settings. At 1000 instances, ME-Mamba reduced GPU memory usage by 39.4\% compared to CCL; at 20,000 instances, the reduction reached 62.2\%, demonstrating its superior efficiency in processing large whole-slide images. Second, we analyzed the FLOPs required by each method to quantify computational efficiency (Fig. \ref{figure3} (b)). ME-Mamba achieved the lowest FLOPs across all three instance counts, confirming its high efficiency. At 10,000 instances, ME-Mamba reduced computational operations by 65.2\% compared to CMTA, highlighting the advantage of the Mamba architecture for multimodal fusion tasks, consistent with its lower memory footprint.

To further demonstrate the robustness of our method, we also extracted pathological image features using two authoritative foundational models, UNI\cite{62} and CONCH\cite{63}, and compared the results with top multimodal methods. Our approach remained highly competitive and achieved the best overall performance.

\begin{figure*}[t!]
    \centering
    \includegraphics[width=\textwidth]{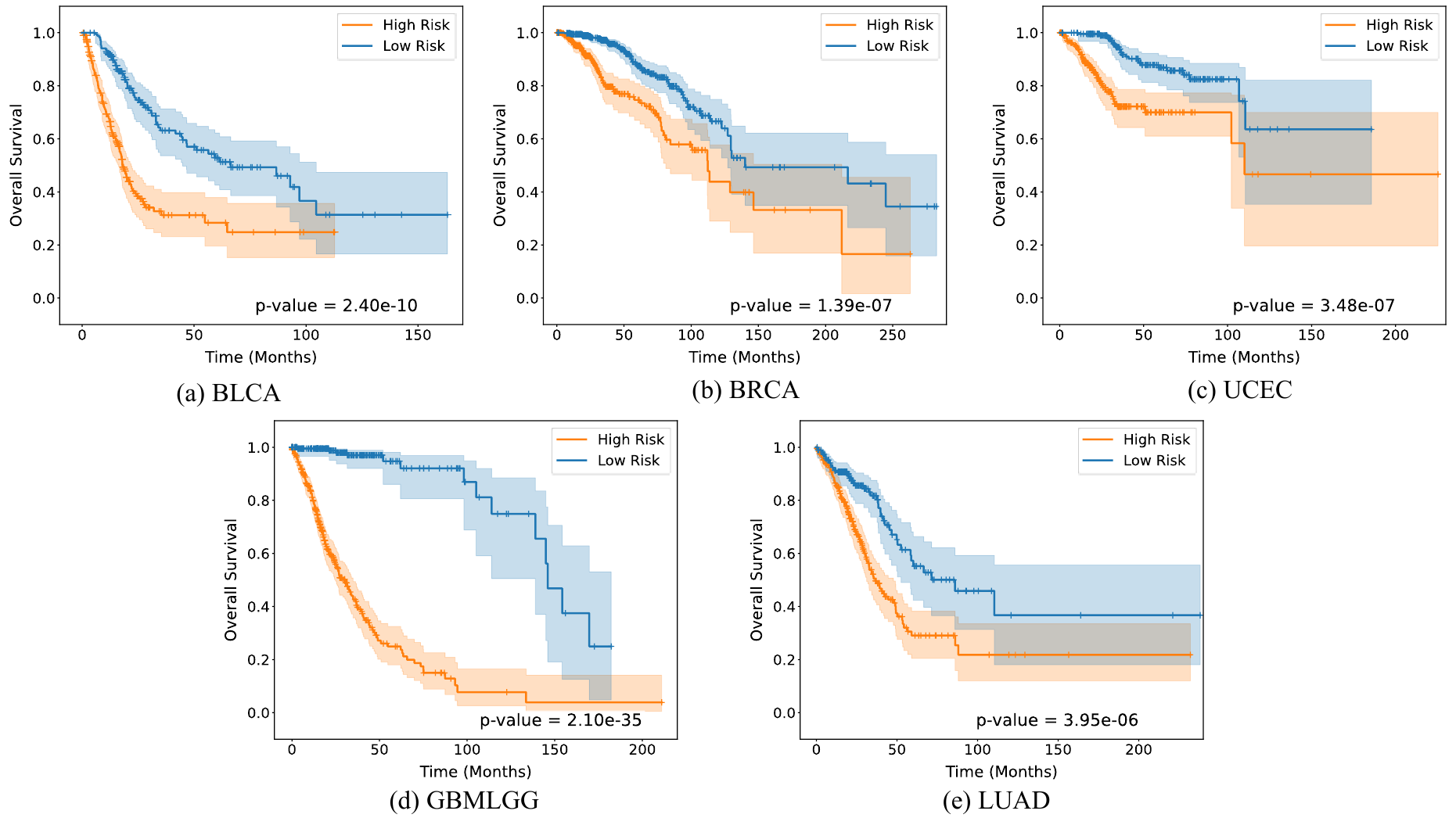}
    \caption{Kaplan-Meier Analysis on five TCGA datasets, where patient stratification of low risk (blue) and high risk (red) are presented. Shaded areas refer to the confidence intervals. p-value < 0.05 means significant statistical differences in two groups, and lower p-value is better.}
    \label{figure4}
\end{figure*}

\subsection{Kaplan-Meier analysis}
To further validate the effectiveness of ME-Mamba for survival analysis, we employ Kaplan–Meier analysis to visualize time-to-event outcomes across all patients. The Kaplan–Meier estimator is a non-parametric statistical method used to estimate survival functions and analyze time-to-event data. Specifically, we first predict a risk score for each patient using our model. Patients are then divided into high-risk and low-risk groups based on the median risk score. Finally, we visualize the survival events using Kaplan–Meier curves, as shown in Fig. \ref{figure4}. A log-rank test was applied to assess the statistical significance of the difference between the high-risk (red curve) and low-risk (blue curve) groups. The resulting p-values across all five datasets were significantly below 0.05, indicating strong discriminatory power of our model in survival analysis. We also compared the p-values obtained by our method with those of other state-of-the-art approaches, as summarized in Table \ref{table2}. Our method achieved the lowest p-values on the BRCA, UCEC, and GBMLGG datasets, while demonstrating highly competitive performance on BLCA and LUAD. These results illustrate the generalizability and robustness of our model across multiple cancer types, suggesting its potential to enhance clinical decision-making and cancer research through reliable survival predictions.

\begin{table}[t]
\centering
\caption{Comparison between p-values of Kaplan-Meier analysis. Bolded red are used to denote the best values.}
\resizebox{\textwidth}{!}{%
\begin{tabular}{cccccc}
\hline
\textbf{Methods} & \textbf{BLCA}                            & \textbf{BRCA}                            & \textbf{UCEC}                            & \textbf{GBMLGG}                          & \textbf{LUAD}                            \\ \hline
SNN              & 1.4e-6                                 & 1.5e-2                                 & 9.8e-4                                 & 1.1e-27                                 & 1.1e-3                                 \\
TransMIL         & 5.6e-2                                 & 2.4e-3                                 & 9.8e-2                                 & 2.5e-25                                 & 3.8e-2                                 \\
MCAT             & 7.8e-2                                 & 7.5e-3                                 & 4.8e-3                                 & 1.6e-19                                 & 6.9e-5                                 \\
SurvPath         & 8.2e-6                                 & 1.4e-3                                 & 1.4e-5                                 & 1.0e-29                                 & 2.2e-4                                 \\
MOTCat           & 2.9e-7                                 & 4.9e-4                                 & 3.8e-7                                 & 3.4e-30                                 & 1.1e-5                                 \\
CMTA             & 2.0e-8                                 & 3.1e-3                                 & 1.7e-3                                 & 1.8e-33                                 & {\color[HTML]{EE0000} \textbf{9.6e-7}} \\
CCL              & {\color[HTML]{EE0000} \textbf{5.7e-11}} & 2.2e-7                                 & 3.6e-4                                 & 9.8e-32                                 & 1.1e-4                                 \\ \hline
\textbf{Ours}    & 2.4e-10                                 & {\color[HTML]{EE0000} \textbf{1.4e-7}} & {\color[HTML]{EE0000} \textbf{3.5e-7}} & {\color[HTML]{EE0000} \textbf{3.1e-35}} & 4.0e-6                                 \\ \hline
\end{tabular}}
\label{table2}
\end{table}

\begin{figure*}[t!]
    \centering
    \includegraphics[width=\textwidth]{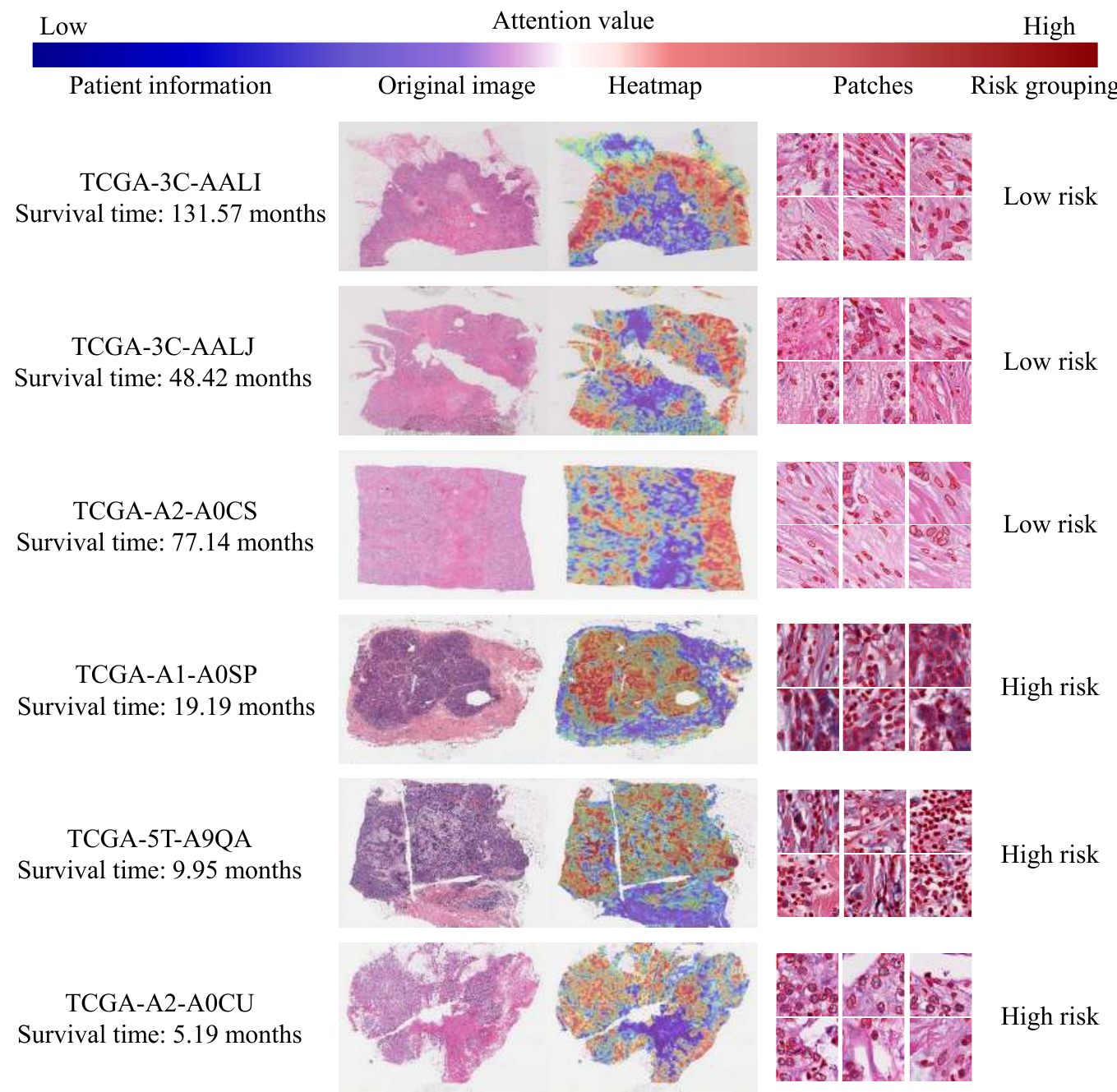}
    \caption{Visualization of heatmaps of pathological slides.}
    \label{figure5}
\end{figure*}

\subsection{Interpretation analysis}
\subsubsection{Visualization of heatmaps of WSIs}
We conducted an interpretability analysis of the Pathology Expert within the multi-expert system to further demonstrate its superior performance, as illustrated in Fig. \ref{figure5}. We use attention weights to create heatmaps on WISs to highlight the representative pathological patches. Specifically, the model assigns a score to each image patch based on its contribution to the final prediction—higher scores indicate greater importance.We normalize the weights between 0 and 1 (i.e., blue to red), select the top six highest-scoring patches for detailed visualization, and segment the nuclei. As shown in Fig. \ref{figure5}, patches with higher weights exhibit similar visual characteristics, such as variably sized nuclei, nuclear atypia, and abnormal nuclear-to-cytoplasmic ratios, indicating that the model can adaptively focus on tumor regions to assist pathologists in diagnosis. Furthermore, high-weight patches from patients with different risk levels display distinct morphological patterns. For example, in the low-risk case “TCGA-3C-AALI” (survival time: 131 months), patches show low-grade pleomorphism. In contrast, the high-risk case “TCGA-5T-A9QA” (survival time: 9.95 months) exhibits patches with severely crowded nuclei. These heatmap visualizations demonstrate that our model can effectively localize discriminative regions in WSIs, which is critical for accurate survival prediction.

\begin{figure*}[t!]
    \centering
    \includegraphics[width=\textwidth]{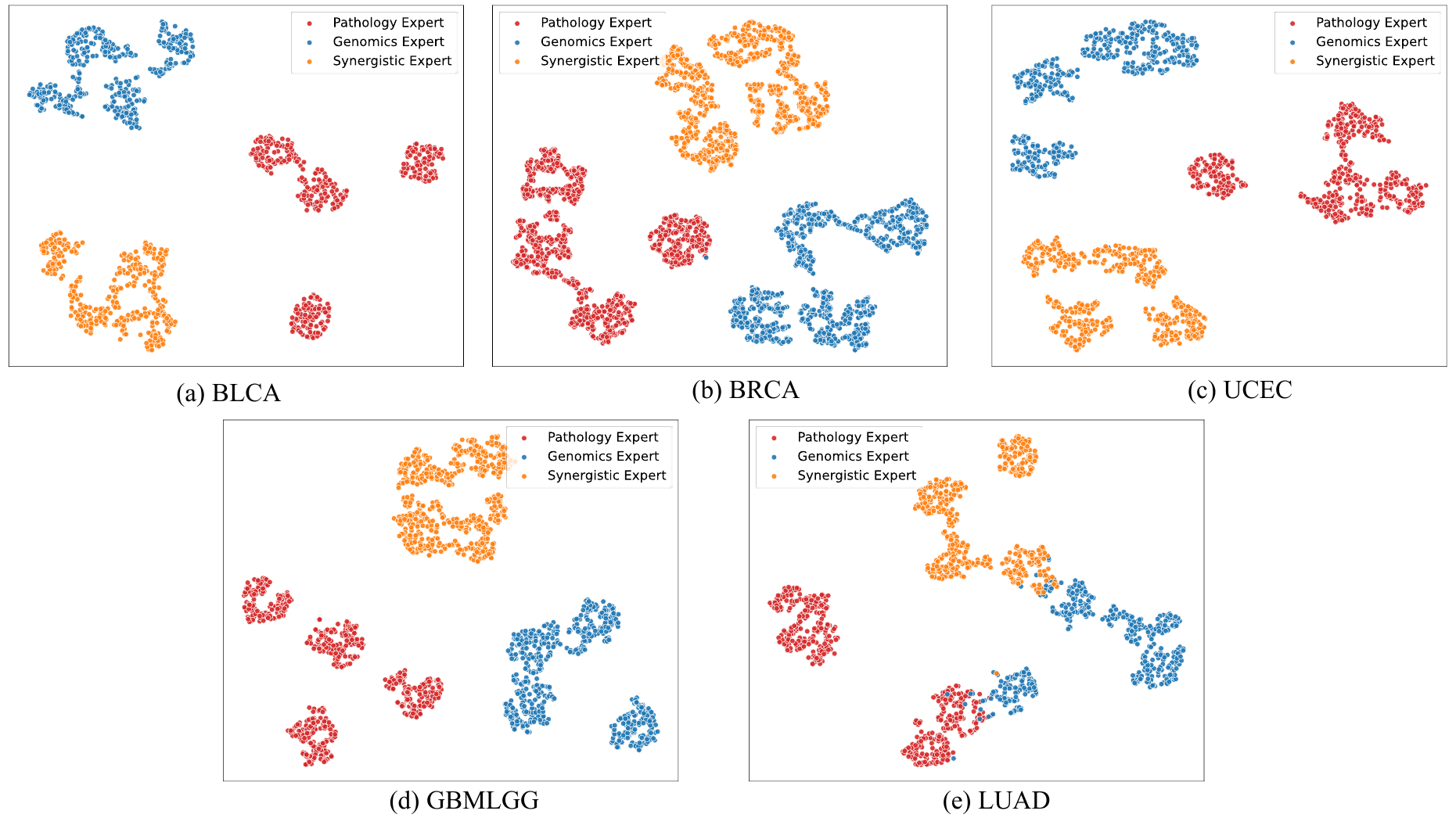}
    \caption{T-SNE visualization of our method on five TCGA datasets.}
    \label{figure6}
\end{figure*}

\subsubsection{T-SNE visualization of features}
We employed T-SNE\cite{64} to visualize the feature distributions output by Pathology Expert, Genomics Expert, and Synergistic Expert. T-SNE minimizes the distances between similar data points and maximizes those between dissimilar points in a low-dimensional space. As shown in Fig. \ref{figure6}, points of different colors represent the feature distributions from different experts. The clear separation and minimal overlap among the feature sets strongly indicate the complementary roles of each expert module in contributing to the final prediction. Within each expert, the feature points are further clustered into approximately four groups, which aligns with the division of ground truth survival time into four discrete intervals. This clustering behavior confirms that each expert in our system effectively captures discriminative patterns relevant to survival prediction.

\begin{table}[t]
\centering
\caption{Ablation results of Synergistic Expert.}
\resizebox{\textwidth}{!}{%
\begin{tabular}{ccccccc}
\hline
\textbf{Method}          & \textbf{BLCA}                                 & \textbf{BRCA}                                 & \textbf{UCEC}                                 & \textbf{GBMLGG}                               & \textbf{LUAD}                                 & \textbf{overall}                       \\ \hline
w/o   Synergistic Expert & 0.6683±0.0289                                 & 0.6483±0.0324                                 & 0.6236±0.0191                                 & 0.8341±0.0143                                 & 0.6577±0.0288                                 & 0.6864                                 \\
All components           & {\color[HTML]{EE0000} \textbf{0.6993±0.0270}} & {\color[HTML]{EE0000} \textbf{0.6909±0.0378}} & {\color[HTML]{EE0000} \textbf{0.7063±0.0248}} & {\color[HTML]{EE0000} \textbf{0.8669±0.0147}} & {\color[HTML]{EE0000} \textbf{0.7014±0.0236}} & {\color[HTML]{EE0000} \textbf{0.7330}} \\ \hline
\end{tabular}}
\label{table3}
\end{table}

\begin{figure*}[t!]
    \centering
    \includegraphics[width=\textwidth]{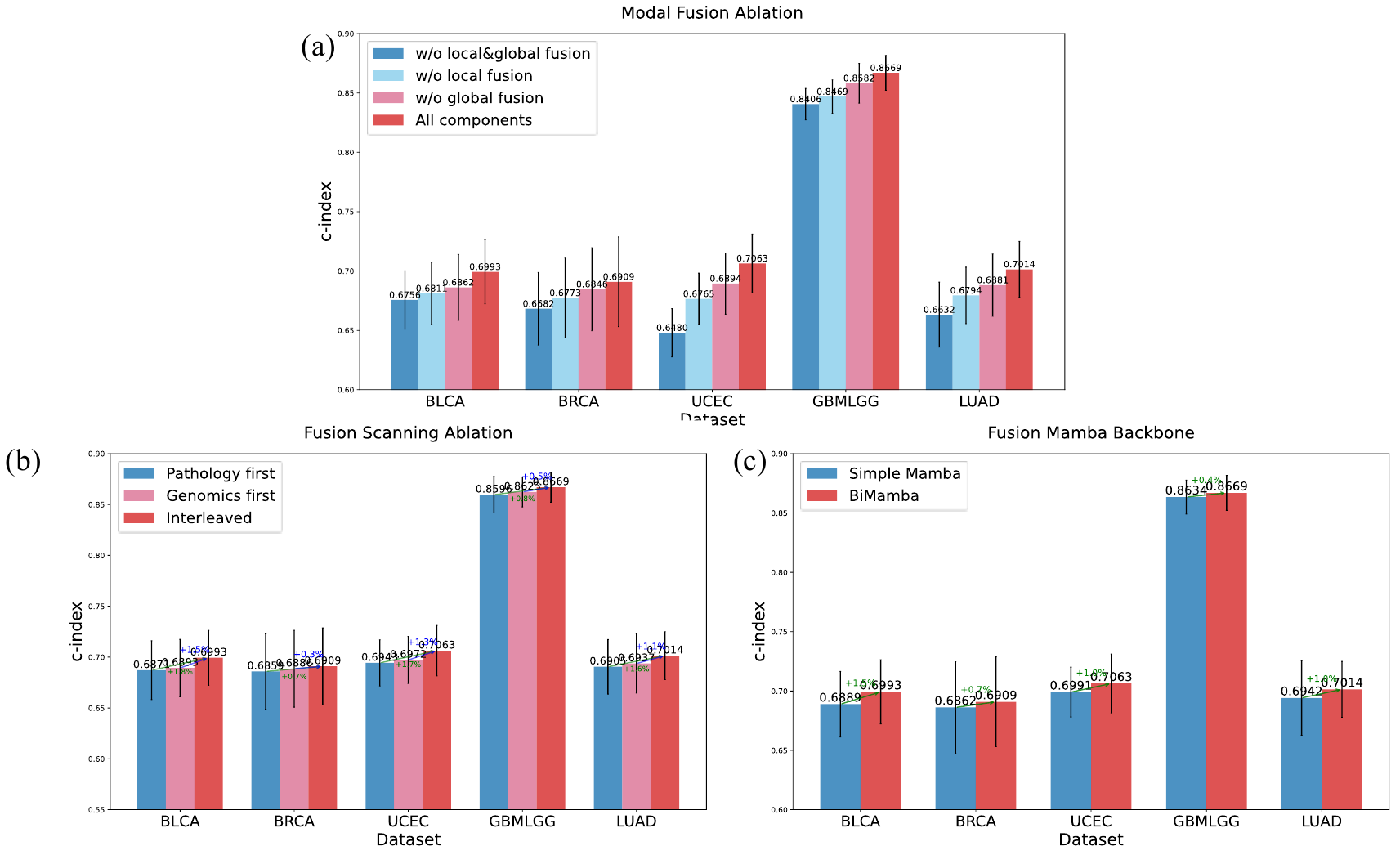}
    \caption{Ablation results of three stages of Synergistic Expert. (a) Ablation of feature fusion stage. (b) Ablation of feature scanning stage. (c) Ablation of feature encoding stage.}
    \label{figure7}
\end{figure*}

\subsection{Ablation Study}
\subsubsection{Effectiveness of Synergistic Expert}
To validate the effectiveness of the proposed Synergistic Expert, we conducted ablation experiments. First, we removed the Synergistic Expert entirely and evaluated the model’s performance. As shown in Table \ref{table3}, the performance declined significantly across all five datasets: the C-index decreased by 4.4\% on BLCA, 6.2\% on BRCA, 11.7\% on UCEC, 3.8\% on GBMLGG, and 6.2\% on LUAD. These results strongly demonstrate the importance of the Synergistic Expert in cross-modal fusion.

Next, we retained the Synergistic Expert and divided its multimodal feature processing into three stages: feature fusion, feature scanning, and feature encoding.

In stage of feature fusion, we evaluated the contributions of both local cross-modal fusion and global cross-modal fusion. The results are shown in Fig. \ref{figure7} (a). Removing either component led to a performance drop. For example, on the UCEC dataset, removing local fusion reduced the C-index by 4.2\%, while removing global fusion resulted in a 2.4\% decrease. Notably, the model performed better when local fusion was retained and global fusion was removed, compared to the opposite scenario. This can be attributed to the fact that local fusion explicitly aligns tokens between modalities via Optimal Transport, while global fusion implicitly matches feature distributions using MMD.

In stage of feature scanning, we compared the proposed interleaved scanning of both modalities with two alternative scanning strategies: (1) Scanning the pathological feature sequence first, followed by the genomic sequence. (2) Scanning the genomic feature sequence first, followed by the pathological sequence. As illustrated in Fig. \ref{figure7} (b), both alternative strategies underperformed the interleaved approach across all datasets. This is because sequential scanning causes the model to focus predominantly on one modality at a time, limiting effective fusion. Moreover, scanning genomic features first led to a smaller performance decline than scanning pathological features first, aligning with the observation that unimodal genomic methods generally outperform unimodal pathological methods, likely due to the stronger correlation of genomic data with survival outcomes.

In stage of feature encoding, we compared the proposed BiMamba backbone with a simple mamba model. Results in Fig. \ref{figure7} (c) show that BiMamba achieved higher c-index values, owing to its bidirectional modeling capability, which processes both forward and backward sequence contexts and enhances performance in dense prediction tasks.

\begin{table}[t]
\centering
\caption{Ablation results of scanning strategies of Pathology Expert and Genomics Expert.}
\resizebox{\textwidth}{!}{%
\begin{tabular}{ccccccccc}
\hline
\textbf{OS} & \textbf{TS}          & \textbf{AS}          & \textbf{BLCA}                                 & \textbf{BRCA}                                 & \textbf{UCEC}                                 & \textbf{GBMLGG}                               & \textbf{LUAD}                                 & \textbf{overall}                       \\ \hline
\checkmark           & \multicolumn{1}{l}{} & \multicolumn{1}{l}{} & 0.6815±0.0252                                 & 0.678±0.0348                                  & 0.6877±0.0186                                 & 0.8503±0.0172                                 & 0.6876±0.0296                                 & 0.7170                                 \\
\checkmark           & \checkmark                    & \multicolumn{1}{l}{} & 0.6879±0.0262                                 & 0.6856±0.0365                                 & 0.6971±0.0225                                 & 0.8612±0.0144                                 & 0.6932±0.0264                                 & 0.7250                                 \\
\checkmark           & \multicolumn{1}{l}{} & \checkmark                    & 0.6866±0.0267                                 & 0.6862±0.0369                                 & 0.6954±0.0250                                 & 0.8634±0.0151                                 & 0.6958±0.0247                                 & 0.7255                                 \\ \hline
\checkmark           & \checkmark                    & \checkmark                    & {\color[HTML]{EE0000} \textbf{0.6993±0.0270}} & {\color[HTML]{EE0000} \textbf{0.6909±0.0378}} & {\color[HTML]{EE0000} \textbf{0.7063±0.0248}} & {\color[HTML]{EE0000} \textbf{0.8669±0.0147}} & {\color[HTML]{EE0000} \textbf{0.7014±0.0236}} & {\color[HTML]{EE0000} \textbf{0.7330}} \\ \hline
\end{tabular}}
\label{table4}
\end{table}

\subsubsection{Effectiveness of Pathology Expert and Genomics Expert}
Our Pathology Expert and Genomics Expert Mamba modules incorporate three scanning strategies: the original scan, the transposed scan (together forming the conventional scanning strategies), and the proposed attention-guided scan. To evaluate the effectiveness of these multiple scanning mechanisms, we designed three experimental settings for comparison: using only the original scan, Using both the original and transposed scans and using both the original and attention-guided scans. The results are summarized in Table \ref{table4}. Experimental results demonstrate that employing two or more scanning strategies consistently improves performance compared to using the original scan alone. This enhancement is attributed to the ability of multiple scanning strategies to more comprehensively capture relationships among all instances. Furthermore, we observed that combining the original scan with the proposed attention-guided scan yields higher performance than combining the original and transposed scans. This is because sorting instances by attention score allows the model to focus more on discriminative instances that are highly correlated with survival outcomes. However, using only two scanning strategies still does not achieve optimal performance. The best results are attained when all three strategies are combined. This integrated approach enables the model to simultaneously capture both discriminative instance-level information relevant to survival and global contextual relationships, while the original scan compensates for feature details that might be overlooked by the other two strategies.

\section{Conclusion}
\label{sec105}
In this paper, we present Multi-Expert Mamba (ME-Mamba), a pioneering system for multimodal survival analysis that synergizes pathology images and genomics data through three specialized experts. This represents a significant advancement in applying the Mamba architecture to multimodal survival analysis tasks. In our system, the Pathology Expert and Genomics Expert employ an attention-guided scanning mechanism to extract discriminative features from gigapixel WSIs and high-dimensional genomic data, explicitly capturing critical instance-level information while preserving global context. The Synergistic Expert combines optimal transport (OT)-based local token fusion and maximum mean discrepancy (MMD)-based global distribution matching to comprehensively model cross-modal interactions. The resulting representations are further refined through a BiMamba backbone to produce enriched multimodal features. Extensive experiments on five TCGA datasets validate the state-of-the-art performance, high computational efficiency, and strong clinical interpretability of ME-Mamba. Given its powerful and generalizable performance, the proposed model can be extended to integrate more data modalities in the future, paving the way for its adaptation to more complex tasks involving diverse data types.

\bibliography{mybibfile}

\end{document}